\documentclass{article}

\usepackage{PRIMEarxiv}

\usepackage[utf8]{inputenc} 
\usepackage[T1]{fontenc}    
\usepackage{hyperref}       
\usepackage{url}            
\usepackage{booktabs}       
\usepackage{amsfonts}       
\usepackage{nicefrac}       
\usepackage{microtype}      
\usepackage{lipsum}
\usepackage{fancyhdr}       
\usepackage{graphicx}       
\graphicspath{{media/}}     

\title{Perception-Driven Bias Detection in Machine Learning via Crowdsourced Visual Judgment
}

\author{
  Chirudeep Tupakula, Rittika Shamsuddin \\
  Department of Computer Science \\
  Oklahoma State University \\
  Stillwater\\
  \texttt{\{Chirudeep.tupakula, r.shamsuddin\}@okstate.edu} \\
}

\begin{document}
\maketitle

\begin{abstract}
Machine learning systems are increasingly deployed in high-stakes domains, yet they remain vulnerable to bias—systematic disparities that disproportionately impact specific demographic groups. Traditional bias detection methods often depend on access to sensitive labels or rely on rigid fairness metrics, limiting their applicability in real-world settings. This paper introduces a novel, perception-driven framework for bias detection that leverages crowdsourced human judgment. Inspired by reCAPTCHA and other crowd-powered systems, we present a lightweight web platform that displays stripped-down visualizations of numeric data (e.g., salary distributions across demographic clusters) and collects binary judgments on group similarity. We explore how users’ visual perception—shaped by layout, spacing, and question phrasing—can signal potential disparities. User feedback is aggregated to flag data segments as biased, which are then validated through statistical tests and machine learning cross-evaluations. Our findings show that perceptual signals from non-expert users reliably correlate with known bias cases, suggesting that visual intuition can serve as a powerful, scalable proxy for fairness auditing. This approach offers a label-efficient, interpretable alternative to conventional fairness diagnostics, paving the way toward human-aligned, crowdsourced bias detection pipelines.
\end{abstract}


\section{Introduction}
Machine learning (ML) systems are increasingly deployed in high-stakes domains such as healthcare, finance, criminal justice, and education. While these systems offer the promise of scalable, data-driven decision-making, they are also susceptible to bias—systematic disparities in model behavior that can disproportionately impact certain demographic groups. Bias can arise from a range of sources, including unrepresentative training data, flawed proxies, historical inequalities, or opaque modeling decisions. Left unchecked, such biases can perpetuate or amplify existing societal injustices.

Traditional approaches to bias detection typically rely on structured group labels (e.g., gender, race) and formal fairness metrics (e.g., demographic parity, equalized odds). However, in many practical settings, these sensitive attributes are unavailable, incomplete, or ethically problematic to use. Additionally, fairness metrics often operate post hoc, offering limited guidance during early stages of model development or data collection.

In this paper, we propose a complementary approach to fairness auditing—one that centers on \textbf{human visual perception}. Inspired by CAPTCHAs and crowdsourced labeling systems, we investigate whether human intuition can serve as an early warning signal for potential bias in numerical datasets. Specifically, we present users with synthetic visualizations of grouped data (e.g., employee salaries segmented by age or experience) and ask them to make rapid, intuitive judgments about whether the groups appear visually similar or different. These responses are collected at scale through a lightweight web-based interface and analyzed to identify cases where perceived group disparity may suggest an underlying bias.

This perception-driven method does not rely on sensitive labels or predefined fairness definitions. Instead, it leverages the human ability to detect patterns—such as skew, clustering, or separation—that might otherwise go unnoticed by algorithmic fairness checks. Our framework also explores how question phrasing influences user responses and how aggregated feedback can be used to guide further statistical analysis and model retraining.

Our key contribution is a novel, label-efficient, and interpretable framework for bias detection that incorporates human judgment into the ML auditing pipeline. By positioning human perception as a scalable “sensor” for fairness, we lay the groundwork for developing more inclusive and transparent AI systems, especially in label-scarce or early-stage development contexts. Furthermore, by aggregating perception-based responses over many visualizations, we can create a labeled dataset of human fairness judgments. These can then be used to train machine learning models that mimic human perception, enabling automated pre-screening of new datasets for potential bias without requiring access to sensitive demographic attributes.

\section{Related Work}

Bias and Fairness in Machine Learning
\cite{dwork2012fairness,hardt2016equality,feldman2015certifying,zafar2017fairness,kleinberg2017inherent,verma2018fairness,chouldechova2017fair,barocas2016big,suresh2019framework,mehrabi2021survey,buolamwini2018gender} \\

Human-in-the-Loop Fairness Auditing and Stakeholder Perspectives
\cite{holstein2019improving,madaio2020co,veale2018fairness,binns2018percentage,saxena2019how, mitchell2019model}\\

Visualization and Perceptual Methods for Bias Detection
\cite{cabrera2019fairvis,ahn2020fairsight}\\

Crowdsourcing, Labeling Pipelines, and Human Feedback for Alignment
\cite{vonahn2008recaptcha,deng2009imagenet,christiano2017deep,ouyang2022training}

\subsection{Bias in Machine Learning}

A growing body of research has documented the risks of bias in machine learning systems, especially when deployed in socially sensitive domains such as healthcare, criminal justice, and employment. Bias can emerge from sample imbalance, flawed measurement proxies, historical discrimination encoded in training labels, or structural constraints in model design. High-profile failures—such as the COMPAS recidivism algorithm disproportionately flagging Black defendants as high-risk, or healthcare algorithms underestimating needs of Black patients due to cost-based proxies—underscore the real-world harm such bias can inflict.

While several fairness metrics have been proposed (e.g., demographic parity, equalized odds, predictive parity), these typically rely on access to accurate and complete group membership labels. In practice, such labels may be missing, sensitive, or ethically fraught to use. This motivates the need for alternative detection frameworks that do not depend on predefined demographic attributes.

\subsubsection{Types of Bias in Machine Learning}

Bias in machine learning systems can stem from various sources across the data pipeline. Key types include:

\begin{itemize}
    \item Sample Bias: Arises when training data fails to represent the diversity of the real-world population. Underrepresentation of certain groups can lead to systematic performance degradation for those groups.

    \item Measurement Bias: Introduced through proxy variables or flawed data collection methods (e.g., using healthcare costs as a proxy for health needs), which may reflect systemic inequities.

    \item Historical Bias: Occurs when seemingly accurate data reflects past injustices—such as biased hiring or policing practices—and carries them forward into algorithmic decisions.

    \item Label Bias: Emerges when the ground-truth labels themselves are subjective or inconsistently applied, often due to human annotators’ biases.

    \item Algorithmic Bias: Results from model architecture, loss functions, or optimization strategies that unintentionally amplify disparities, even when data is balanced.

    \item  Proxy Bias: When non-sensitive attributes (like zip code or browser type) correlate with protected attributes (e.g., race, gender), leading to indirect discrimination.
\end{itemize}

These biases often interact, making them difficult to detect and mitigate using purely technical means. Our approach sidesteps some of these challenges by relying on perceptual signals rather than demographic metadata.

\subsubsection{Bias Detection and Fairness Metrics}
Traditional methods for detecting bias involve computing fairness metrics such as demographic parity, equalized odds, or disparate impact. While useful, these metrics often rely on known sensitive attributes and assume clearly defined group boundaries—assumptions that do not always hold in practice. Furthermore, these metrics provide quantitative post hoc analysis, but offer limited visibility into how bias is perceived by end-users.

\subsection{Human-in-the-Loop Fairness Auditing}

Recent work has emphasized the importance of human involvement in fairness-aware AI development. Human-in-the-loop frameworks introduce stakeholder input at various stages—from data labeling to model evaluation—to identify and mitigate harms that automated systems might overlook. Participatory design and expert-in-the-loop evaluations allow humans to guide model behavior, especially in complex or value-laden contexts.

However, most human-in-the-loop systems require either domain expertise or extensive user effort, limiting their scalability. Our work takes a more lightweight approach, using rapid, binary perceptual judgments from non-experts to flag potential disparities in visualized data—making the process more scalable and accessible.

\subsection{Visualization and Perceptual Bias}

Visualization has long been a tool for interpreting machine learning behavior. Techniques such as t-SNE and UMAP are used to visualize high-dimensional data embeddings and to surface potential representational disparities. Saliency maps, attention heatmaps, and other interpretability tools help explain model predictions, but are primarily intended for expert audiences.

Our approach departs from these in two ways: (1) it uses deliberately simplified visualizations (e.g., dot plots) that require no technical background to interpret, and (2) it uses perception not as a diagnostic tool for understanding models, but as a primary signal for detecting group-level bias in data representations.

\subsection{Crowdsourcing and CAPTCHA-Inspired Labeling}

Crowdsourcing has played a central role in AI development, from the creation of ImageNet to human feedback used in training large language models. reCAPTCHA systems, for instance, successfully redirected human effort into useful labeling tasks like image classification and book digitization. Reinforcement Learning with Human Feedback (RLHF) now powers the alignment of generative models with user preferences.

We draw inspiration from this lineage to propose \textbf{Fairness CAPTCHAs}—small, perceptual tasks designed to elicit human feedback about potential disparities between groups. Rather than labeling correctness, our goal is to collect intuitive, perception-based signals of unfairness. This novel repurposing of crowdsourcing establishes human perception as a viable resource for fairness auditing in structured data.

\section{Methodology}

Our proposed framework combines human perceptual judgment with statistical analysis and machine learning to detect and eventually automate the identification of bias in structured datasets. The methodology is divided into two main phases: (1) Perception-based bias flagging via crowdsourcing, and (2) Model calibration and automation using labeled human feedback. The full pipeline — from perception capture to automated ML response — is illustrated in Figure \ref{fig:overview}.

\subsection{Data Preparation and Visualization Design}

\begin{figure}
  \centering
  \includegraphics[width = 10cm, height = 14cm]{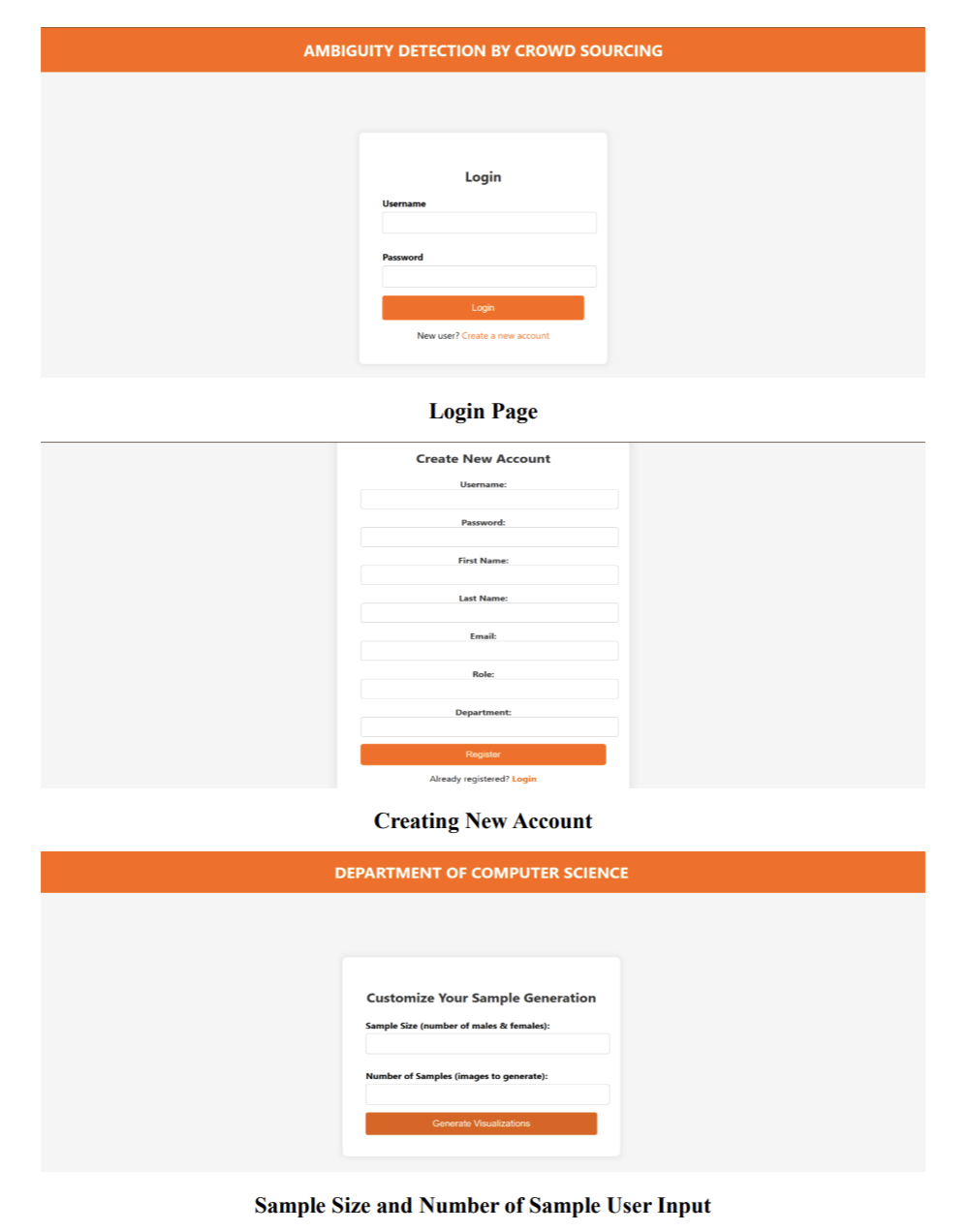}
  \caption{ Web-based user interface for login, registration, and customization of sample generation. The platform allows users to set group sizes and the number of visualizations per session.}
  \label{fig:app}
\end{figure}

We begin by partitioning the original dataset into subgroups based on features such as age and experience. Each subgroup pair (e.g., high age–high experience vs. low age–low experience) is then visualized as a 2D scatter plot using color-coded dots. These visualizations are deliberately minimal — omitting axis labels, units, and numerical scales — to abstract away domain-specific context and focus user attention on relative group structure, density, and clustering patterns. This design choice serves two purposes: (1) to test whether users can perceive statistical disparities without formal cues, and (2) to reduce the influence of prior beliefs, social bias, or label-driven anchoring. By stripping the visuals of sensitive or loaded information, we aim to elicit more neutral, perception-based assessments of fairness.

These visualizations are generated dynamically and served to users through a web application (see Figure \ref{fig:app}) built with Flask and PostgreSQL. The application is deployed via a Flask backend with PostgreSQL for session logging and response tracking. The platform is device-agnostic and designed to support fast, lightweight interaction.

\subsection{Phase 1: Crowdsourced Visual Judgment}
\begin{figure}
  \centering
  \includegraphics[width = 9cm, height = 12cm]{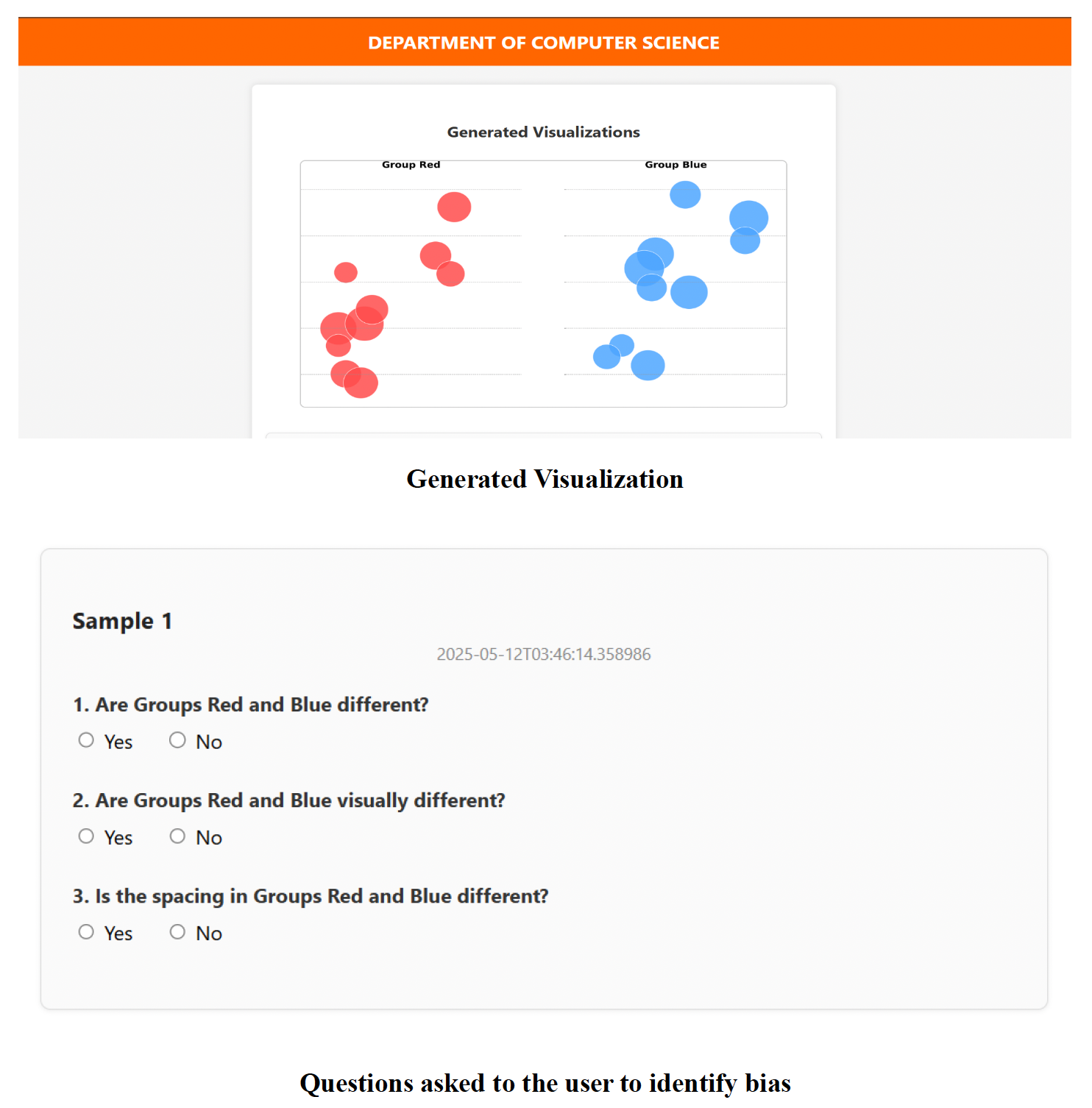}
  \caption{  Interface showing a generated visualization alongside three versions of the perception question. The application randomizes image-question pairs and logs binary (yes/no) responses. Users think they are playing a game of "Spot the Difference".}
  \label{fig:survey}
\end{figure}

Users are presented with a sequence of visualizations and asked binary yes/no questions such as, “Do these two groups look visually similar?” or “Do you observe a noticeable difference?” (see Figure \ref{fig:survey}). The question phrasing is systematically varied across users to test how linguistic framing influences perception.

Each user response is stored with metadata including timestamp, phrasing version, and optionally device type or response latency. We aggregate responses across multiple users for each image-question pair to assess perceived fairness. Optional metadata such as response time and user confidence scores are stored to enable future weighting or uncertainty modeling.

\subsection{Statistical Validation of Perception-Flagged Bias}

\begin{figure}
  \centering
  \includegraphics[width = 6cm, height = 8cm]{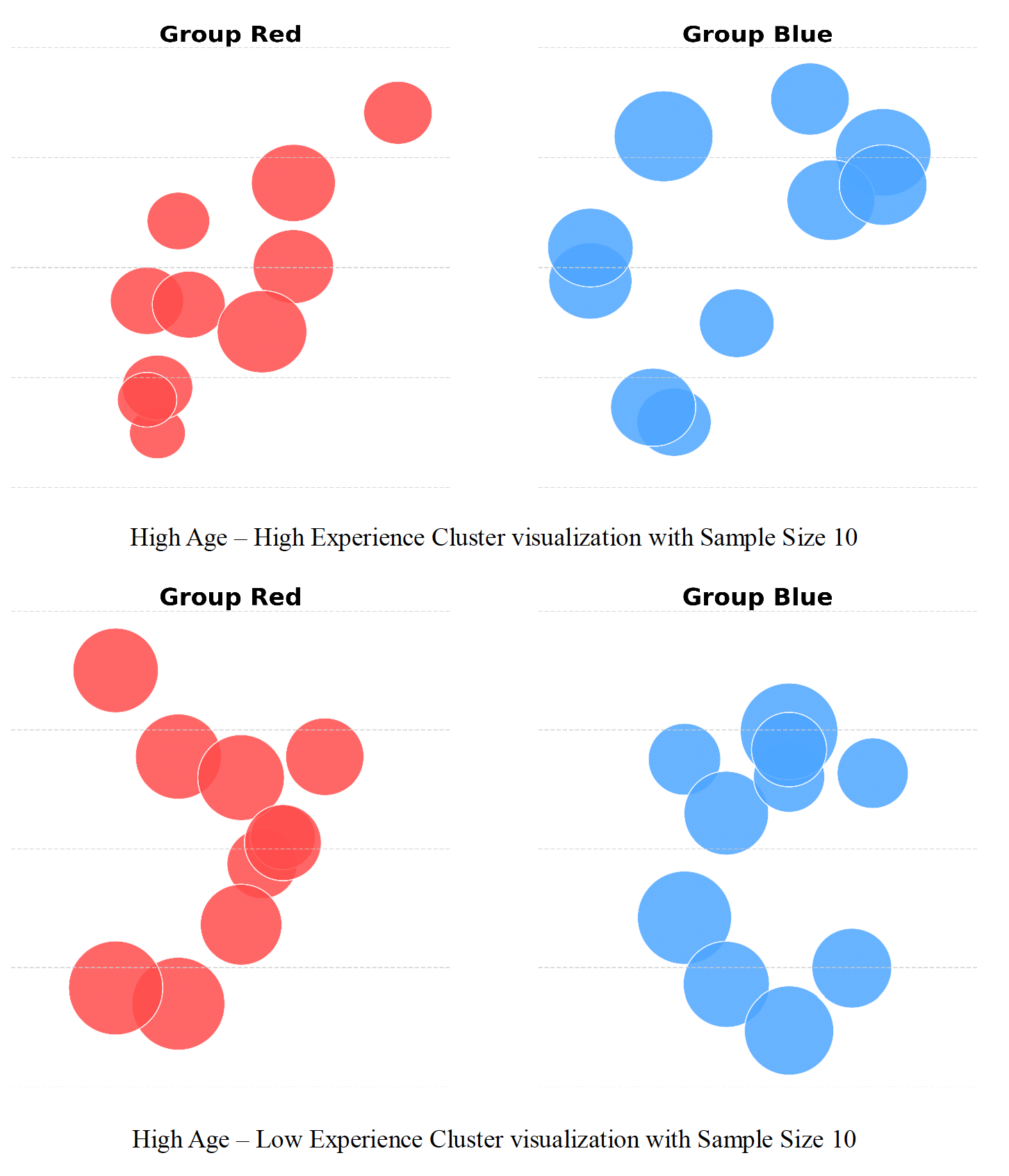}
   \includegraphics[width = 6cm, height = 8cm]{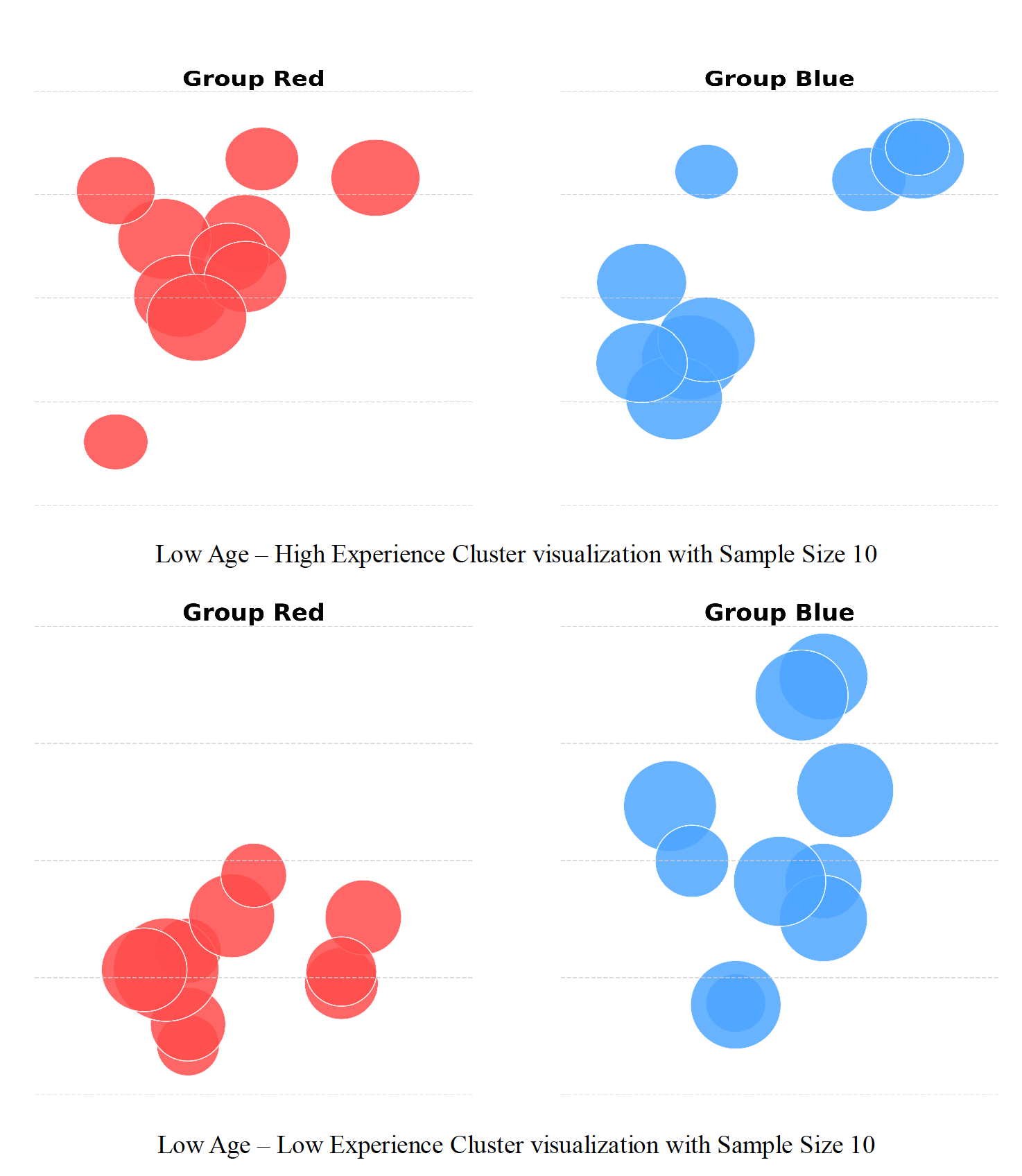}
  \caption{ Sample cluster visualizations presented to users. Groups are color-coded and plotted without axis labels to focus attention on spatial patterns and minimize cognitive bias. Users think they are playing a game of "Spot the Difference".}
  \label{fig:dataslice}
\end{figure}

While human visual perception provides an intuitive signal for identifying potential disparities, it is inherently subjective. To validate and calibrate user-flagged cases with greater rigor, we apply standard data science methodologies to establish a ground truth. Specifically, we use two-sample t-tests to statistically compare the means of sensitive variables (e.g., salary) between the visualized subgroups. This approach helps determine whether perceived differences are supported by statistically significant disparities in the underlying data (see Figure \ref{fig:dataslice}).

A case is considered calibrated if it meets two criteria: (1) a majority of users perceive a disparity in the visualization (e.g., >X\% of responses signal "bias"), and (2) the statistical test returns a p-value below a predefined threshold (typically 
p<0.05). This dual-filtering mechanism—combining crowd perception with statistical testing—ensures that flagged instances are grounded in empirical evidence, reducing the risk of false positives and enhancing interpretability for downstream analysis.

\subsection{Phase 2: Training Bias-Predicting Models}
Calibrated perception-based labels are used to train machine learning models that learn to mimic human fairness judgment. We experiment with classifiers (e.g., decision trees, SVMs) that take visual or numerical encodings of dataset slices as input and output binary fairness labels (bias/no bias).

The goal is to automate the early detection of bias in future data slices without relying on sensitive attributes. Once trained, these models can act as a screening tool to flag new cases where group disparities may exist, prompting further review or model refinement.

We experiment with standard supervised learning algorithms such as decision tree classifiers and support vector machines (SVMs).The input features include:

\begin{enumerate}
\item Encoded representations of the visualized data slices (e.g., group-wise means, variances, distribution statistics),
\item Cluster-specific context (e.g., age/experience groupings),
\item Structural metrics derived from plot layouts.

\end{enumerate}

These models learn to predict whether a given data slice would likely be perceived as biased by a majority of users. This enables scalable, real-time bias screening of new dataset segments or model outputs, even in the absence of sensitive attributes or labels.

\subsection{Feedback Loop and Future Automation}
\begin{figure}
  \centering
  \includegraphics[width = 14cm, height = 10cm]{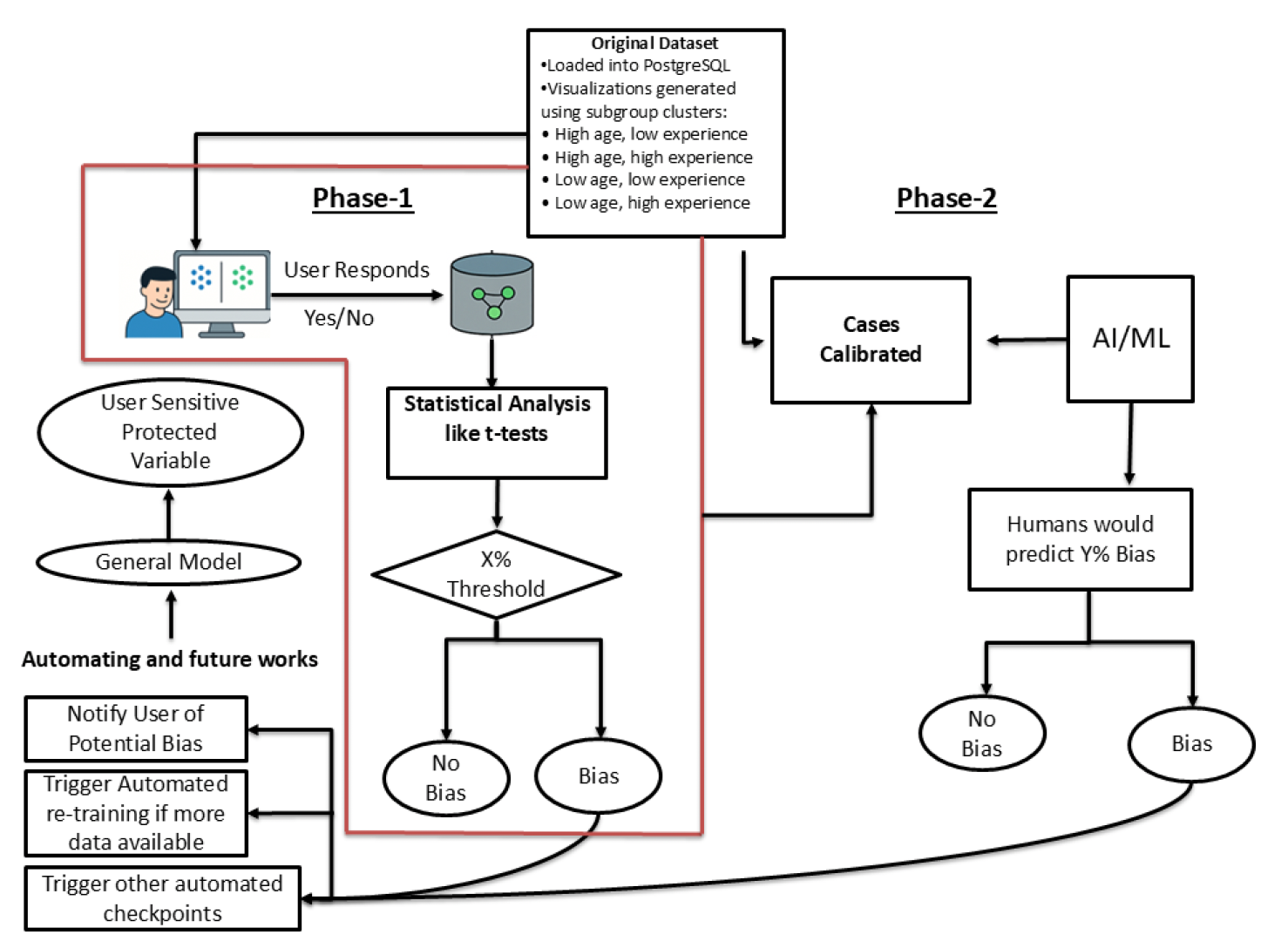}
  \caption{ Overview of the proposed two-phase framework for bias detection. Phase 1 captures user-perceived disparities through visual judgment and flags cases for statistical testing. In Phase 2, calibrated labels are used to train ML models that mimic human bias perception, enabling automated fairness screening and future re-training triggers.}
  \label{fig:overview}
\end{figure}

As shown in Figure \ref{fig:overview}, the system is designed to support an automated feedback loop. Once bias is detected and validated, the framework can trigger alerts, suggest retraining the general model on balanced data, or activate other checkpoints in the ML pipeline. Over time, as more user feedback is collected, the bias-predicting models improve in reliability, closing the loop from human perception to ML-driven fairness audits.

\subsubsection{ Automated System Responses}
To close the loop, the system includes a set of automated responses that are triggered when bias is flagged by either human perception or the trained ML model:
\begin{enumerate}
\item User Notification: Inform stakeholders (e.g., data scientists or ML engineers) of flagged data segments or model behavior that warrants further inspection.
\item Retraining Trigger: If additional training data is available (e.g., more balanced or representative samples), the system can initiate re-training or fine-tuning of the affected model.
\item Audit Checkpoints: Integration with automated fairness auditing tools or CI/CD pipelines for model deployment, enabling bias flags to pause deployment or log compliance risks.

\item Downstream Bias Impact Evaluation: To assess whether observed data bias leads to biased model behavior, the system also supports training subgroup-specific models (e.g., trained separately on male and female subsets). These models are cross-evaluated on the opposite group (e.g., train on males, test on females), and performance metrics such as Mean Squared Error (MSE) are compared. A significant drop in cross-group performance suggests that data bias has led to model bias.

\item Prediction Distribution Analysis: Additionally, the system can compare prediction distributions across groups to detect skew or unfair calibration. For instance, if a model consistently predicts lower outcomes for one group, this may indicate embedded bias in either the training data or the model’s learned behavior.

\end{enumerate}

\subsubsection{Split-and-Cross-Evaluation}
As an additional checkpoint, we implement subgroup-based model training and evaluation. Specifically, we partition the dataset by a protected attribute (e.g., gender) and train separate models for each subgroup (e.g., one model on male data, another on female data). These models are then evaluated both in-group and cross-group. For example, a model trained on male data is tested on female instances, and vice versa. A significant increase in prediction error (e.g., MSE) during cross-group testing may indicate that the model has learned specific patterns that fail to generalize — signaling the presence of algorithmic bias and/or a distributional difference across groups with respect to the target variable (e.g., salary). This evaluation helps disentangle whether disparities flagged by human perception are rooted in the data, the model, or both, and provides a deeper diagnostic signal for fairness auditing.

We experiment with models such as Support Vector Regression and Decision Tree Regressors for both subgroup modeling and perception-aligned bias prediction.

These evaluation methods help distinguish between \textbf{bias in data representation} and \textbf{bias in model behavior}, enabling more targeted mitigation strategies. For example, cross-group model testing and prediction distribution analysis can reveal whether observed disparities in the data propagate into performance gaps or skewed outputs. When bias is confirmed, the system can take actionable steps: notifying users, triggering retraining, or initiating fairness audits. As more perception-labeled examples are collected, the bias-detection model itself can be periodically retrained, increasing its precision and generalizability. These feedback mechanisms ensure that the framework is not only diagnostic but also \textbf{extensible and intervention-ready}, enabling timely and structured responses to emerging fairness issues.

\section{ Experiments and Results}

\subsection{ Pilot Deployment and Participants}
To evaluate the effectiveness of our perception-driven bias detection framework, we deployed a prototype web application to a pilot group of participants. The cohort included university students and colleagues from varied educational backgrounds. No prior knowledge of statistics, fairness, or machine learning was required to participate.

Participants accessed the system via desktop and mobile browsers, registered accounts, and interacted with a randomized set of visualizations and survey questions. Figures \ref{fig:app} and \ref{fig:survey} shows the application’s user interface and setup screens, including options to customize sample sizes and session parameters.

\subsection{Visualizations and Task Design}
Participants were presented with a sequence of synthetic 2D scatter plots, each displaying grouped data represented by color-coded dots (e.g., red vs. blue). These visualizations were carefully designed to vary along several perceptual dimensions, including the degree of separation between groups, internal clustering or density, and overall spatial balance or skew. By controlling these visual properties, the experiment aimed to simulate a range of subtle or overt disparities that might arise in real-world data representations.

Each visualization was paired with a binary yes/no question, prompting users to assess whether the two groups appeared visually similar or different. To examine the impact of linguistic framing on perception, the question phrasing was varied systematically across sessions—for example, “Do these two groups look visually similar?” versus “Do you observe a noticeable difference between the groups?”

Figure \ref{fig:dataslice} illustrates representative examples from four distinct clusters that were evaluated during the study: High Age–High Experience, High Age–Low Experience, Low Age–High Experience, and Low Age–Low Experience. These clusters reflect realistic combinations of demographic attributes and allow for analysis of both perceptual and statistical disparities across a diverse range of subgroup configurations.

\subsection{Collected Metrics and Behavioral Patterns}
During each session, the application logged several types of data for every user response. These included the binary yes/no answer to the visual fairness question, the session metadata such as device type, and the exact phrasing of the question presented. Optional metrics, like response time and user-reported background (e.g., educational domain or professional field), were also collected to support future subgroup analysis and model weighting strategies.

Analysis of the collected data revealed several noteworthy patterns. Visualizations that exhibited stronger asymmetry, imbalance, or internal clustering were significantly more likely to be flagged as biased by participants. Moreover, question phrasing played a measurable role in influencing perception: users were more inclined to identify bias when the prompt emphasized potential disparity (e.g., “Do you notice a difference?”) than when it framed the task around similarity (e.g., “Do the groups look similar?”). Importantly, responses were found to be highly consistent across repeated exposures and different users, suggesting that visual perception of group-level disparity can serve as a reliable and reproducible signal for fairness concerns.

\subsection{Statistical Validation of Group Disparities}

To test whether user-flagged images also reflected statistically significant differences in the underlying data, we conducted two-sample t-tests on group-level salary distributions across the four demographic clusters. The analysis compared mean salaries between male and female subgroups within each cluster and reported the corresponding p-values.

\begin{table}[h]
\centering
\begin{tabular}{|l|c|c|c|c|}
\hline
\textbf{Cluster} & \textbf{Male Mean} & \textbf{Female Mean} & \textbf{p-value} & \textbf{Statistically Significant?} \\
\hline
High\_High & 12583.10 & 15222.83 & 0.03557 & Yes \\
High\_Low  & 14565.93 & 15318.95 & 0.32890 & No \\
Low\_High  & 16190.84 & 19865.15 & 0.10992 & No \\
Low\_Low   & 13577.32 & 14324.95 & 0.28779 & No \\
\hline
\end{tabular}
\caption{Two-sample t-test results comparing male and female salary means across clusters.}
\label{tab:ttests}
\end{table}

Only the \textit{High Age – High Experience} cluster showed a statistically significant difference between male and female salary distributions (\(p = 0.03557\)). This result aligned with the high user-flag frequency for this visualization, thereby confirming the validity of perception-driven signals and supporting the effectiveness of crowd-based bias detection.

\subsection{Machine Learning Model Validation}

To assess whether flagged disparities also affected downstream model behavior, we trained subgroup-specific regression models and evaluated their performance using cross-group testing. Specifically, models were trained separately on male-only and female-only data within each cluster, and their predictive accuracy was compared using Mean Squared Error (MSE). We used both Support Vector Regression (SVR) and Decision Tree Regressors implemented in \texttt{scikit-learn}.

\begin{table}[h]
\centering
\begin{tabular}{|l|l|c|c|c|c|}
\hline
\textbf{Cluster} & \textbf{Model} & \textbf{M$\rightarrow$M} & \textbf{M$\rightarrow$F} & \textbf{F$\rightarrow$F} & \textbf{F$\rightarrow$M} \\
\hline
High\_High & MSE & 12583.10 & 15222.83 & 12472.62 & 12515.86 \\
High\_Low  & MSE & 14565.93 & 15318.95 & 14888.89 & 16681.17 \\
Low\_High  & MSE & 16190.84 & 19865.15 & 17344.54 & 18524.78 \\
Low\_Low   & MSE & 13577.32 & 14324.95 & 13595.36 & 14179.25 \\
\hline
\end{tabular}
\caption{Cross-group model performance (MSE) for male- and female-trained models across demographic clusters.}
\label{tab:crossmse}
\end{table}

In all clusters, we observed a degradation in cross-group performance. For instance, models trained on male data consistently yielded higher prediction error when tested on female data, and vice versa. This pattern was especially pronounced in the \textit{High Age – High Experience} cluster, mirroring the earlier statistical and perception-based findings. These results suggest the presence of algorithmic bias and/or underlying differences in group-level distributions with respect to the target variable (e.g., salary), reinforcing the value of layered bias diagnostics.

\subsection{Summary of Findings of the Pilot Study}

The results from our pilot study demonstrate that human users can reliably perceive disparities in synthetic data visualizations—even in the absence of sensitive group labels. However, for these perceptual signals to be useful in fairness auditing, they must be calibrated against statistical ground truth, as subjective impressions alone can be inconsistent or misleading. In our analysis, several user-flagged cases, particularly the High Age – High Experience cluster, aligned closely with statistically significant differences in salary distributions, supporting the validity of crowd-sourced perception as a preliminary bias signal.

Moreover, cross-group evaluation of subgroup-specific models revealed that some visually identified disparities were associated with real and quantifiable prediction gaps, especially when tested across demographic boundaries. This correspondence between perception and model performance reinforces the effectiveness of our layered auditing pipeline, which combines human intuition with data science methodologies.

The study also revealed that both question phrasing and visual structure had a measurable impact on user judgments, highlighting the role of cognitive framing in fairness perception. While initial trends were consistent, these framing effects warrant further controlled testing to better understand their influence and ensure robustness across larger and more diverse participant pools.

These findings are based on a limited, early-stage pilot study intended primarily to test the feasibility of the system and gather preliminary insights. While the results are promising and suggest that perception-based feedback can serve as a reliable proxy for bias detection, more rigorous, large-scale, and scientifically controlled testing is needed to validate these observations across diverse populations and real-world datasets. Nonetheless, the initial alignment between user perception, statistical validation, and model behavior provides a strong foundation for future work. We are actively expanding the system for broader deployment, with plans to incorporate more diverse datasets, gamified interfaces, and larger participant pools to refine and validate the methodology further.

\section{Discussion and Implications}

\subsection{Interpretation of Results}

Our pilot results suggest that human visual perception, even from non-expert users, can effectively surface patterns of potential bias in structured data — particularly when disparities are visible in group clustering, density, or spatial separation. These intuitive signals were most reliable when triangulated with statistical significance and machine learning model performance, underscoring the power of a layered fairness auditing approach. The strong alignment in the High Age – High Experience cluster across perception, data, and model metrics highlights the potential of this hybrid methodology for real-world bias detection.

Importantly, these results offer preliminary validation — enough to justify further exploration — of using perceptual judgments not as a replacement for formal fairness metrics, but as a low-cost, early-stage signal for bias detection. Such perceptual signals can help prioritize areas for deeper analysis before investing in more resource-intensive statistical or algorithmic evaluations. This framework is particularly valuable in contexts where sensitive labels are unavailable, incomplete, or ethically sensitive, and where conventional fairness metrics may not be feasible due to legal or technical constraints.

\subsection{Limitations and Future Work}

While the findings are encouraging, several limitations must be acknowledged. First, although the underlying dataset is drawn from real-world sources (e.g., Kaggle), the visualizations derived from it are not controlled experiments — meaning we cannot isolate the effects of specific visual features or data properties as precisely as we could with synthetic data. This lack of experimental control makes it difficult to definitively attribute user responses to specific visual cues, motivating future work that begins with fully synthetic, parametrically varied datasets before returning to real-world complexity.

Second, the participant pool was limited in size and diversity, which may have introduced demographic bias into the perception judgments. Broader sampling across cultural and professional backgrounds is necessary to assess the generalizability of the framework.

Third, while we explored multiple question phrasings and visual encodings, these were not systematically counterbalanced or randomized across all users. As a result, the interaction between cognitive framing, visual layout, and user background remains an open research question.

Finally, human perception itself is inherently subjective. While this subjectivity is core to our approach, it introduces variability that must be carefully managed through aggregation and calibration against statistical ground truth.

\subsection{Implications for Fairness Auditing and ML Practice}
Despite these limitations, our findings suggest several valuable contributions to the ML fairness community. First, the system offers a scalable, label-efficient auditing mechanism that can surface bias even in data-limited scenarios. Second, by capturing human intuition directly, the approach bridges the gap between technical audits and public perception, which is increasingly critical in domains like healthcare, hiring, and automated decision-making. Third, the system enables the training of bias-predicting models that mimic human perception, opening the door to automated, perception-aligned fairness screening.

This research invites a broader rethinking of how humans can contribute to ethical AI pipelines—not just as annotators or auditors, but as interpreters of fairness itself.

\section{Conclusion}
This work presents a novel, perception-driven framework for detecting bias in machine learning datasets using crowd-sourced human judgment. By leveraging intuitive visual comparisons of group-level data distributions, our system enables lightweight, label-efficient bias detection that does not depend on sensitive demographic attributes or formal fairness metrics. Our pilot study demonstrates that users can reliably perceive disparities in structured data, and that these perceptions often align with statistical significance and model performance degradation—especially in high-risk clusters.

The proposed two-phase architecture—beginning with human perception and followed by statistical validation and model cross-evaluation—offers a scalable, interpretable pipeline for fairness auditing. It helps bridge the gap between algorithmic diagnostics and human-aligned values, especially in early-stage or label-scarce scenarios.

While preliminary in scope, our findings support the viability of perception-based signals as a proxy for bias and highlight several avenues for future research. These include larger and more diverse deployments, expansion to real-world ML outputs, and development of trained models that can replicate human fairness perception at scale. Ultimately, this work contributes toward a more inclusive, transparent, and interpretable approach to fairness in AI systems—one that centers human intuition alongside statistical rigor.

\bibliographystyle{unsrt}  
\bibliography{references}

\end{document}